\documentclass{article}

\usepackage{neurips_2024}
\usepackage{amsmath,amssymb}
\usepackage{siunitx} 

\usepackage[utf8]{inputenc} 
\usepackage[T1]{fontenc}    
\usepackage{hyperref}       
\usepackage{url}            
\usepackage{booktabs}       
\usepackage{amsfonts}       
\usepackage{nicefrac}       
\usepackage{microtype}      
\usepackage{xcolor}         

\title{Agent-Orchestration in Autonomous Chip Design}

\author{%
  Linyang Li \\
  Nova Silicon\\
  \texttt{lilinyang@nova-silicon.com} \\
}

\begin{document}

\maketitle

\begin{abstract}
  Recent developments in large language models (LLMs) and tool-using agents encourage people to explore the potential of using agents in chip design. The core question is what kind of AI we really need in such a sophisticated industry.
  To this end, we bring the idea of modeling a chip-design superintelligence as an enormous \textit{AI-organization}.
\end{abstract}

\section{Superintelligence Native Chip Design}

The most critical convenience of coding agents is the concept of \textbf{vibe coding}, despite its remaining flaws.
The core change is that we go from measuring speed as a percentage‑of‑human, to measuring speed as a percentage‑of‑model.
That is, the only valuable game-changing chip design technique is an AI system that works autonomously.

With such a strength, we can build chips that can serve various AI models, algorithms, and applications, enabling these people who are seeking custom chips for their business, such as different training or inference scenarios, different model structures, and different agent workflows.

\section{How to construct such superintelligence?}

\subsection{Modeling Action Space}

Considering chip design or any experimental science procedures in one test-run,  we consider \textbf{action space} as the set of all feasible actions that an agent or optimization algorithm can take.
Within optimization frameworks:
Optimization variables are treated as actions;
Each set of variable values is regarded as a single action;
The collection of all admissible variable value assignments forms the Action Space.

More frequently, we face black-box optimization problems, that is, a set of actions (design parameters) is fed into a black-box simulator or physical experiment, followed by output of performance metrics. No analytical gradients are available, and evaluation can only be performed via sampling.
Analog circuit sizing is a canonical black-box optimization scenario~\citep{jones1998efficient,krylov2023learning}.

In analog circuit design optimization for blocks such as operational amplifiers, LDOs, and PLLs. Tunable parameters correspond to dimensions of MOS transistors, where $W$ denotes transistor width and $L$ denotes transistor length. Circuit performance metrics (objective functions) include gain, bandwidth, phase margin, power consumption, input offset voltage, output swing, etc. Constraints cover minimum area, maximum power budget, process design limits and saturation region requirements.

Within this scenario:
\begin{itemize}
    \item $\text{Action Space} =$ the set of all valid combinations of transistor $W$ and $L$ values;
    \item Each selection of a $(W,L)$ combination corresponds to one action executed by the optimization algorithm.
\end{itemize}
\vspace{1mm}

\paragraph{Dimensionality of the Action Space}
\paragraph{Definition of Dimensionality}
The dimensionality of the action space equals the number of independent tunable variables:
\[
\dim(\text{Action Space}) = N
\]
\textbf{Example}: A circuit containing 5 MOS transistors, with both $W$ and $L$ optimized for each device. The independent variables are $W_1,L_1,W_2,L_2,\dots,W_5,L_5$, forming a 10-dimensional action space.

Action spaces fall into two major categories:
\begin{itemize}
    \item \textbf{Continuous Action Space}: Variables take real-number values. Transistor $W$ and $L$ belong to this category. Widely adopted black-box optimization algorithms include Bayesian optimization, random search, and reinforcement learning algorithms~\citep{snoek2012practical}.
    \item \textbf{Discrete Action Space}: Variables can only take a finite set of discrete values (e.g., predefined device sizing bins). This formulation is rarely used natively in circuit optimization.
\end{itemize}

\paragraph{$\boldsymbol{\Rightarrow}$ Critical Note: Curse of Dimensionality}
As the dimensionality of the action space increases, the number of samples required to uniformly cover the entire space grows exponentially. A 10-dimensional circuit optimization problem is substantially harder than a 3-dimensional counterpart. Grid search or exhaustive enumeration becomes infeasible in high-dimensional settings. This is the core advantage of black-box optimizers such as Bayesian optimization over brute-force search.

\paragraph{Search Boundaries of the Action Space (Box Constraints)}
An action space defines not only the number of variables but also the valid upper and lower bounds for each variable:
\[
\boldsymbol{a} = [a_1,a_2,\dots,a_N],\quad a_i \in [low_i,\;high_i]
\]
Returning to analog transistor sizing, variable ranges are strictly bounded by manufacturing processes and design rules. For a single transistor pair $(W,L)$:
\[
L_i \in [L_{\text{min}},\;L_{\text{max}}]
\]
$L_{\text{min}}$ denotes the minimum gate length specified by the PDK (e.g., $0.028\,\mu\text{m}$ for a 28 nm process). $L_{\text{max}}$ sets the upper limit; excessively long gate lengths introduce extra parasitics and are restricted by area specifications defined by designers.
\[
W_i \in [W_{\text{min}},\;W_{\text{max}}]
\]
$W_{\text{min}}$ is the minimum allowable transistor width per process rules. $W_{\text{max}}$ is the upper bound, as overly large width leads to excessive area overhead and layout routing difficulties.

\paragraph{Feasible Domain vs. Hyperrectangular Search Space}
Most optimization frameworks adopt a hyperrectangular search domain by imposing independent upper and lower bounds on each variable:
\[
low_1\le a_1\le high_1,\; low_2\le a_2\le high_2,\;\dots
\]
This region constitutes the raw action space. However, analog circuits contain implicit coupled constraints: for instance, differential pair matching enforces $W_1/L_1 = W_2/L_2$, and similar matching rules apply to current mirrors. Under such conditions, numerous points inside the raw hyperrectangular action space violate physical design constraints. The true feasible region forms a low-dimensional manifold embedded within the original space.

Two mainstream engineering strategies exist for black-box optimization:
\begin{enumerate}
    \item Define a loosely bounded action space. After sampling, verify constraints via circuit simulation and apply penalty terms to infeasible samples.
    \item Leverage prior design constraints to reduce dimensionality and shrink the action space, lowering search difficulty (the preferred approach for circuit optimization).
\end{enumerate}

\paragraph{End-to-End Workflow for Black-Box Optimization in Circuit Sizing}
The complete optimization pipeline for analog transistor sizing is outlined below:
\begin{enumerate}
    \item Define the action space: select transistors to be optimized, specify upper and lower bounds for each $W$ and $L$, and construct an $N$-dimensional continuous search domain.
    \item The optimization algorithm samples an action vector $\boldsymbol{a}$ within the action space.
    \item Pass parameter set $\boldsymbol{a}$ into the EDA simulator, treated as a black-box model with no analytical gradients available.
    \item Run circuit simulation to extract performance metrics for objective evaluation and constraint checking.
    \item The algorithm uses historical sampling data to guide the selection of sampling locations in subsequent iterations within the action space.
    \item Repeat iterations until a transistor sizing combination satisfying all specifications is found.
\end{enumerate}

Careful construction of the action space is critical for efficient black-box optimization. An overly broad search range drastically reduces sampling efficiency, with many samples falling into invalid regions. An excessively narrow range risks convergence to local optima and excludes globally optimal sizing solutions. Introducing redundant independent variables exacerbates the curse of dimensionality. In engineering practice for automated analog circuit optimization, properly pruning and refining the action space often yields greater performance gains than switching to a different optimization algorithm.

In Chip design, transistor sizing is a well-bounded subproblem within chip design that can be formally defined with relative ease. 
By contrast, other tasks across the chip design flow (topology synthesis, placement and routing, architecture exploration, circuit specification refinement, system-level tradeoff analysis, etc.) generally suffer from ambiguous problem definitions and modeling challenges.

\subsection{Language Modeling for Reasoning}

Language Modeling (LM) is a machine learning paradigm that models the probability distribution over sequences. Mathematically, an LM learns the joint distribution \(P(t_1,t_2,...,t_n)\) of a sequence \(s=[t_1,t_2,...,t_n]\), factorized via the chain rule of conditional probability:
\(P(t_1,\dots,t_n)=\prod_{i=1}^n P(t_i|t_1,\dots,t_{i-1})\)
In its original narrow definition, language modeling predicts the next token in natural language sentences.
In the modern generalized sense, Transformer-based Large Language Models (LLMs) are trained on broad text corpora and can condition their outputs on extensive contextual information~\citep{vaswani2017attention}.

The core properties of language modeling is that:

(1) Beyond pure numerical fitting; capable of explicit semantic reasoning.

(2) Embeds rich prior knowledge during pre-training (theoretical equations, design methodologies, device physics).

(3) Can be prompted (or harnessed) to produce intermediate reasoning steps via Chain-of-Thought (CoT) prompting~\citep{wei2022chain}.

Recall analog transistor sizing: tunable \(W,L\) form the Action Space, and SPICE simulation acts as the black-box evaluator.
Fundamental limitations of classical black-box optimizers (Bayesian Optimization, etc.):

(1) No physical semantic awareness:

The optimizer only observes a vector of floating-point numbers \([W_1,L_1,W_2,L_2...]\). It does not understand that these values correspond to transistors, transconductance \(g_m\), \(g_m/I_D\) operating region constraints, or analytical relations between gain and \(g_m\). Search is purely interpolation/extrapolation over sampled data.

(2) Poor generalization:

Experience learned on one circuit topology, specification set or process node cannot transfer. After changing topology or specs, sampling and optimization must restart from scratch. Extrapolation often fails without underlying physical laws.

(3) Inability to adopt human design heuristics:

Human designers follow explicit reasoning: set target \(g_m\) → select bias current via \(g_m/I_D\) methodology → back-calculate proper \(W/L\).
Black-box optimizers cannot execute this logical pipeline and rely on blind search.

(4) Inefficient exploration in high-dimensional Action Space:

Many points within variable bounds violate device operation rules (e.g., transistors in triode region). Infeasibility can only be discovered after costly simulation, wasting computational budget.

To this end, the advantage of modeling LLM-native reasoning:

(1) Knowledge foundation shifts from sampled data to universal physical laws:

All knowledge of classical black-box optimizers is encapsulated within simulation samples collected for the current task. Historical samples become invalid when topology or specifications change.
LLMs embed universal MOS device theory, \(g_m/I_D\) methodology and amplifier principles. This knowledge is general physical law, not tied to a specific circuit instance. When facing new op-amp topologies or new specs, the model can conduct valid reasoning from scratch, enabling zero-shot / few-shot generalization.

(2) Native constraint awareness to prune the Action Space:

During reasoning, the LLM automatically avoids invalid operating regimes: prevents triode-region operation, enforces matching ratios, maintains reasonable \(g_m/I_D\) ranges. Many physically infeasible regions inside the raw action space are eliminated before simulation, mitigating the curse of dimensionality and cutting useless evaluations.

(3) Interpretable optimization and cross-task knowledge transfer:

Classical black-box methods output sizing values without explaining design choices.
LLMs produce explicit rationales: “Gain improvement requires higher input transistor \(g_m\), therefore increase W”. Such design heuristics transfer naturally to other amplifier designs.

(4) Overcome the extrapolation failure limitation of pure black-box methods:

Surrogate models such as Gaussian Processes deliver reliable predictions only near existing samples; confidence degrades rapidly in unsampled regions.
Guided by analytical physics, LLMs support physically grounded extrapolation into unexplored regions, discovering high-quality designs inaccessible to conventional blind search.

\subsection{Orchestration of Autonomous Agents}

Circuit sizing is a one-man, one-day job.
That is, sizing task constitutes a continuous reasoning task, which naturally lends itself to an agent-based workflow. Nevertheless, when targeting more complex engineering circuits, the internal agent logic, decision branches, interactions with domain knowledge, and multi-objective trade-off constraints grow drastically. Such sophisticated agent systems essentially follow an object-oriented organizational paradigm, rather than a simple procedural structure with linear or predefined branches.
Conventional development workflows widely adopted in engineering are largely procedural: they predefine fixed steps, execution sequences, and node dependencies, and are only suitable for standardized tasks with fully enumerable execution paths. When applied to build organization-level AI agent systems, they reveal fundamental limitations. Autonomous agents feature dynamic decision-making, non-exhaustible conditional branches, self-planning, on-demand tool invocation, multi-agent collaboration and dynamic knowledge interaction — behaviors that cannot be fully bounded by static flow graphs. Static workflows lack capabilities including object encapsulation, message passing, independent state management and dynamic behavioral extension. Consequently, they struggle to host large-scale agent architectures with autonomous reasoning, reusable components, and cross-team organizational collaboration.

\subsection{Bottleneck of EDA Simulation}

Contemporary EDA tools are built upon two fundamental components: simulation engines that model objective physical behavior, and extensively manually coded rule-based heuristic algorithms.

Physics-based simulation quantitatively evaluates design performance via numerical solution of device and circuit models with reproducible outputs, though the reproducibility is time-consuming to manifest.

For example, in device simulation and circuit simulation EDAs, device simulation and circuit simulation both aim to predict the physical behavior of real electronic systems through mathematical models, but they operate at different levels of abstraction. Device simulation starts from the physical mechanisms inside semiconductor devices and numerically solves equations such as Poisson's equation and carrier transport equations to predict quantities such as potential, current, and carrier distributions under different geometries, materials, doping profiles, voltages, and temperatures. Circuit simulation, in contrast, abstracts devices such as transistors into compact models and combines them with Kirchhoff's laws to numerically solve circuits containing many interconnected devices, producing circuit-level metrics such as voltage, current, delay, and power. Although these simulators are grounded in physical laws, theoretical equations alone are insufficient to fully describe fabricated hardware. Material parameters, process variations, contact resistance, parasitic effects, temperature dependence, and other non-ideal device behaviors must be extracted, calibrated, and validated against measurements from physical devices and test chips. The construction of a physics-based simulator therefore forms a closed loop of physics → modeling → measurement → calibration → validation: physical laws determine the structure of the model, while interaction with the physical world determines its parameters and verifies whether the simulator can reliably reproduce real device and circuit behavior over the intended operating range. Only through this continuous alignment with physical reality can simulation outputs serve as meaningful predictions of engineering performance rather than merely mathematically self-consistent results.

Heuristic algorithms serve as the backbone of automated flows. These algorithms translate decades of engineering experience, layout constraints, circuit matching principles, and process design rules into hard-coded logic.
After decades of industrial refinement, they are stable, predictable, and well-supported by mature debugging workflows, and have become highly sophisticated, underpinning commercial chip design worldwide.
Traditional EDA tools such as ICC2 and Innovus use explicit design rules to constrain the original combinatorial search space to a feasible region, and then further prune that space through domain knowledge, problem decomposition, and heuristic algorithms. This substantially reduces the portion of the search space that must actually be explored, turning an otherwise computationally intractable optimization problem into one that is practically solvable in engineering workflows.
Recent learning-based chip placement work demonstrates both the promise of learned policies and the importance of tightly integrating them with the EDA flow~\citep{mirhoseini2021graph}. Traditional machine learning, by contrast, primarily performs function fitting on the limited historical data generated by these existing EDA flows. As a result, the training data itself covers only a sparse subset of the original design space—specifically, the subset already filtered and explored by predefined rules and heuristics. For high-frequency cases that lie within the training distribution, ML can learn effective mappings and provide significant acceleration. However, industrial long-tail problems often arise from previously unseen process conditions, constraint combinations, and multivariable interactions, whose combinatorial space grows far faster than the amount of data that can realistically be collected.
More importantly, the long tail in EDA is often not caused by a single variable taking an unusual value, but by a rare combination of otherwise common factors:
\[
A,\ B,\ C,\ D
\]Each of these factors may have appeared individually in the training data, while their joint occurrence,
\[
A\cap B\cap C\cap D,
\]may never have been observed.
Finite datasets cannot exhaustively cover this kind of combinatorial explosion:
\[
|\mathcal X|
\sim
\prod_{i=1}^{n}|X_i|.
\]As the number of variables increases, the number of possible combinations can grow exponentially, while the available dataset does not scale at the same rate. Therefore, purely relying on statistical fitting cannot provide reliable coverage of the full underlying design space, nor can it inherently guarantee correctness in these previously unseen long-tail regions.

Instead of relying on massive manually defined explicit rules, large language models leverage domain knowledge and symbolic reasoning to directly model the physical logic and design methodologies behind circuits, devices, and layouts.
Its long-term evolution unfolds in two stages:
(1) Deploy LLM reasoning to replicate human engineers’ thinking patterns and accomplish design tasks that traditionally demand manual intervention;
(2) Continuously learn from design cases and simulation feedback to gradually reduce, refine, and replace cumbersome legacy rule sets.
A critical distinction from superficial AI plug-in integration: AI-native frameworks do not merely attach an independent model to existing EDA flows for auxiliary usage. They take intelligent reasoning as a foundational capability, fundamentally reducing reliance on extensive handwritten hard-coded rules. Only such a paradigm holds the potential to progressively replace substantial manual design labor and continuously optimize, and eventually supersede traditional rule-based algorithm systems.





\section{Conclusion}

With the opinions we state, we believe that AI will revolutionize future chip design technology.
Providing end-to-end solutions with autonomous AI is an inevitable path, tens to hundreds of times faster than human-in-the-loop chip design.

\end{document}